\documentclass[conference]{IEEEtran}
\IEEEoverridecommandlockouts
\usepackage{amsmath,amsfonts}
\usepackage{threeparttable}
\usepackage{graphicx}
\usepackage{siunitx}
\usepackage{multirow}
\usepackage{proof}
\usepackage{makecell}
\usepackage{array}
\usepackage{amsmath}
\usepackage{amssymb}
\usepackage{epstopdf}
\usepackage{float}
\usepackage{amssymb}
\usepackage{nomencl}
\usepackage{enumerate}
\usepackage{enumitem}
\usepackage[linesnumbered,ruled,vlined]{algorithm2e}
\usepackage{subfigure}
\usepackage{bm}

\usepackage{amsthm}
\usepackage{multirow}
\usepackage{array}
\usepackage[caption=false,font=normalsize,labelfont=sf,textfont=sf]{subfig}
\usepackage{textcomp}
\usepackage{stfloats}
\usepackage{url}
\usepackage{xcolor} 
\usepackage{verbatim}
\usepackage{subcaption} 
\usepackage[font=small]{caption} 
\usepackage[utf8]{inputenc}
\usepackage{graphicx}
\usepackage{cite}
\usepackage{booktabs}
\newcommand{\RNum}[1]{\uppercase\expandafter{\romannumeral #1\relax}}
\usepackage[english]{babel}

\usepackage{wrapfig}
\usepackage{enumitem}
\usepackage[capitalize]{cleveref}
\def\BibTeX{{\rm B\kern-.05em{\sc i\kern-.025em b}\kern-.08em
    T\kern-.1667em\lower.7ex\hbox{E}\kern-.125emX}}
\makeatletter
\newcommand{\linebreakand}{%
  \end{@IEEEauthorhalign}
  \hfill\mbox{}\par
  \mbox{}\hfill\begin{@IEEEauthorhalign}
  
}
\makeatother
\begin{document}

\title{Red-Team Multi-Agent Reinforcement Learning for Emergency Braking Scenario

}

\author{

\IEEEauthorblockN{1\textsuperscript{st} Yinsong Chen}
\IEEEauthorblockA{\textit{School of Mechanical Engineering} \\
\textit{Beijing Institute of Technology}\\
Beijing, China \\
3220240416@bit.edu.cn}
\and
\IEEEauthorblockN{2\textsuperscript{nd} Kaifeng Wang}
\IEEEauthorblockA{\textit{School of Mechanical Engineering} \\
\textit{Beijing Institute of Technology}\\
Beijing, China \\
3120230311@bit.edu.cn}
\and
\IEEEauthorblockN{3\textsuperscript{rd} Xiaoqiang Meng}
\IEEEauthorblockA{\textit{School of Mechanical Engineering} \\
\textit{Beijing Institute of Technology}\\
Beijing, China \\
3120220377@bit.edu.cn}
\and
\IEEEauthorblockN{4\textsuperscript{th} Xueyuan Li}
\IEEEauthorblockA{\textit{School of Mechanical Engineering} \\
\textit{Beijing Institute of Technology}\\
Beijing, China \\
lixueyuan@bit.edu.cn}
\and
\IEEEauthorblockN{5\textsuperscript{th} Zirui Li}
\IEEEauthorblockA{\textit{School of Mechanical Engineering} \\
\textit{Beijing Institute of Technology}\\
Beijing, China \\
z.li@bit.edu.cn}
\and
\IEEEauthorblockN{6\textsuperscript{th} Xin Gao}
\IEEEauthorblockA{\textit{School of Mechanical Engineering} \\
\textit{Beijing Institute of Technology}\\
Beijing, China \\
https://orcid.org/0000-0002-7317-8059}

\linebreakand 
\IEEEauthorblockN{} 
*Co-corresponding authors:  Xueyuan Li and Xin Gao.\\
This work was supported by National Key R\&D Plan of China (Grant No.2024YFB3411301)
}

\maketitle

\begin{abstract}
Current research on decision-making in safety-critical scenarios often relies on inefficient data-driven scenario generation or specific modeling approaches, which fail to capture corner cases in real-world contexts. To address this issue, we propose a Red-Team Multi-Agent Reinforcement Learning framework, where background vehicles with interference capabilities are treated as red-team agents. Through active interference and exploration, red-team vehicles can uncover corner cases outside the data distribution. The framework uses a Constraint Graph Representation Markov Decision Process, ensuring that red-team vehicles comply with safety rules while continuously disrupting the autonomous vehicles (AVs). A policy threat zone model is constructed to quantify the threat posed by red-team vehicles to AVs, inducing more extreme actions to increase the danger level of the scenario. Experimental results show that the proposed framework significantly impacts AVs decision-making safety and generates various corner cases. This method also offers a novel direction for research in safety-critical scenarios.
\end{abstract}


\section{Introduction}
Recent advances in artificial intelligence are accelerating the transition of autonomous vehicles (AVs) from controlled testing environments to open roads \cite{first}. However, the large-scale deployment of autonomous driving is still hindered by the lag in safety-critical scenario research. Current AVs still exhibit a high decision-making failure rate when faced with corner cases, leaving a significant gap compared to the safety and robustness standards required for Level 4 (L4) autonomy. Therefore, advancing research in safety-critical scenarios is essential not only for technical progress but also for the secure and dependable deployment of AVs in real-world contexts.
 
 Contemporary research on safety-critical scenarios primarily adopts two approaches. The first is the data-driven method \cite{biandao}, where real-world driving data is utilized for scenario reconstruction and simulation-based validation. However, due to policy constraints, regulatory barriers, and limited technological maturity, the sparse deployment of AVs leads to inadequate coverage of safety-critical scenarios. Furthermore, such scenarios follow a long-tail distribution in naturalistic driving data, with infrequent occurrences that limit the effectiveness of data-driven models in overcoming cognitive generalization limits \cite{10533445}. The second approach is scenario-specific modeling, where deterministic scenarios are constructed to optimize AVs decision-making. Although this significantly improves the decision-making safety of AVs in specific conditions, it overlooks hidden dangerous maneuvers of background vehicles (BVs), failing to meet the safety requirements for AVs in real-world safety-critical scenarios. As a result, existing researchers fail to address the potential corner cases that may arise in reality, significantly limiting the practical deployment of AVs.

The red-team concept originated in the military, where it serves as a simulated adversary to test a system’s defensive capabilities. The core idea of the red-team is to simulate adversarial tactics and behaviors, compelling the system to expose its weaknesses and vulnerabilities, thereby facilitating improvements before real-world deployment \cite{redteam}. In autonomous driving, the red-team concept involves introducing disruptive behaviors to challenge the decision-making and response capabilities of AVs. In this framework, the red-team is typically composed of BVs, simulating extreme traffic scenarios to compel AVs to respond to these corner cases. Furthermore, the active interference and exploration of red-team vehicles can create more corner cases, filling gaps in real-world driving data.

In summary, we propose a Red-Team Multi-Agent Reinforcement Learning (RMARL) framework, where BVs in traffic scenarios are treated as red-team agents. These agents are trained using reinforcement learning algorithms to develop interference policies, generating adversarial behaviors. Through simulation experiments, red-team vehicles increased the collision rate of AVs from 5\% to 85\%, while generating multiple corner cases in the original scenario. This also demonstrates that the proposed framework provides a novel approach for researching safety-critical scenarios. The main contributions of this paper are as follows:

 \begin{enumerate}[label=(\alph*)]
 \item We propose a RMARL framework that overcomes the limitations of traditional methods, which heavily rely on historical driving data. By utilizing interference from red-team vehicles, the framework actively explores potential high-risk scenarios.
 \item We propose a CGMDP integrated with a Dual-Constrained Graph Proximal Policy Optimization (DC-GPPO) algorithm, which applies hard constraints on the action space to prevent illegal maneuvers by red-team vehicles and soft behavioral constraints to penalize non-disruptive interference.
 \item We propose a PTZ model that quantifies the level of threat red-team vehicles pose to AVs in the scenario, thereby encouraging more dangerous behaviors from red-team agents.
 \item Simulation results reveal that the RMARL framework severely degrades AVs decision-making safety in the scenario, exhibiting a 17-fold surge in collision rates. The inclusion of red-team vehicles also creates various corner cases in the scenario.
 \end{enumerate}

\section{Related Work}
\subsection{Methods based on data-driven}
In recent years, decision-making in safety-critical scenarios has gained substantial traction, with numerous studies employing data-driven methods to enhance the discovery of corner cases. For example, Feng et al. \cite{feng2021intelligent} introduced the Natural–Adversarial Hybrid Environment, which identifies high-impact vehicles and critical time instances using key metrics. Ding et al. \cite{9197145} introduced a conditional multi-trajectory synthesis framework that generates multi-dimensional driving scenarios, meeting safety-critical conditions derived from traffic flow dynamics models, thus offering a comprehensive test set for systemic safety validation. Klischat et al. \cite{klischat2019generating} presented an evolution driven critical scenario generator that guides the search for behavioral boundaries and failure modes of autonomous driving systems by parameterizing scene representations and fitness functions. \cite{ding} decompose traffic scenarios into reusable autoregressive building blocks, and adaptively search for high-risk scenario parameters with the goal of collision.

However, data-driven approaches are inherently limited by the completeness of real-world driving data, due to the rarity of corner cases in natural traffic, such methods struggle to uncover latent high-risk scenarios or differentiate scenarios based on their risk severity. 

\subsection{Methods based on scenario modeling}\label{AA}
Several researchers have undertaken modeling studies of specific safety-critical scenarios. In highway scenarios, Niu et al. \cite{10840090} improved AVs decision-making safety using a risk-aware utility function. Wang et al. \cite{10844066} enhanced AV decision-making safety in rear-end, cutting-in, and lane-changing scenarios by combining traditional trajectory prediction methods with a risk-aware scene encoder. In highway scenarios with emergency vehicle cut-ins, \cite{10774317} used Oriented Bounding Box (OBB) collision detection to assess collision risks, coupled with a state machine model for decision-making, demonstrating improved risk avoidance. 
Fu et al. \cite{9067008} proposed an emergency braking strategy based on the Deep Deterministic Policy Gradient (DDPG) algorithm for lead vehicle lane changes or sudden braking, resulting in a 15\% reduction in collisions. Li et al. \cite{10786483} Developed a cognitive model of driver information accumulation using the drift-diffusion model to examine decision-making processes in rear-end collision scenarios. For lane reduction scenarios, Xu et al. \cite{9338584} combined the Constant Turn Rate and Acceleration model for predicting BV trajectories, optimized decisions with Double Deep Q Networks (DDQN), and applied safety rules for real time action correction.

While focusing exclusively on specific scenarios can markedly improve the decision‑making safety of AVs, it neglects the exploration of corner cases, which impose far more stringent safety requirements than routine conditions. Consequently, we must uncover potential extreme scenarios in traffic environments and use them as a foundation to further enhance AVs safety.

\section{PROBLEM FORMULATION}
\begin{figure}[bt!]
   \centering
     \includegraphics[width=1.13\linewidth]{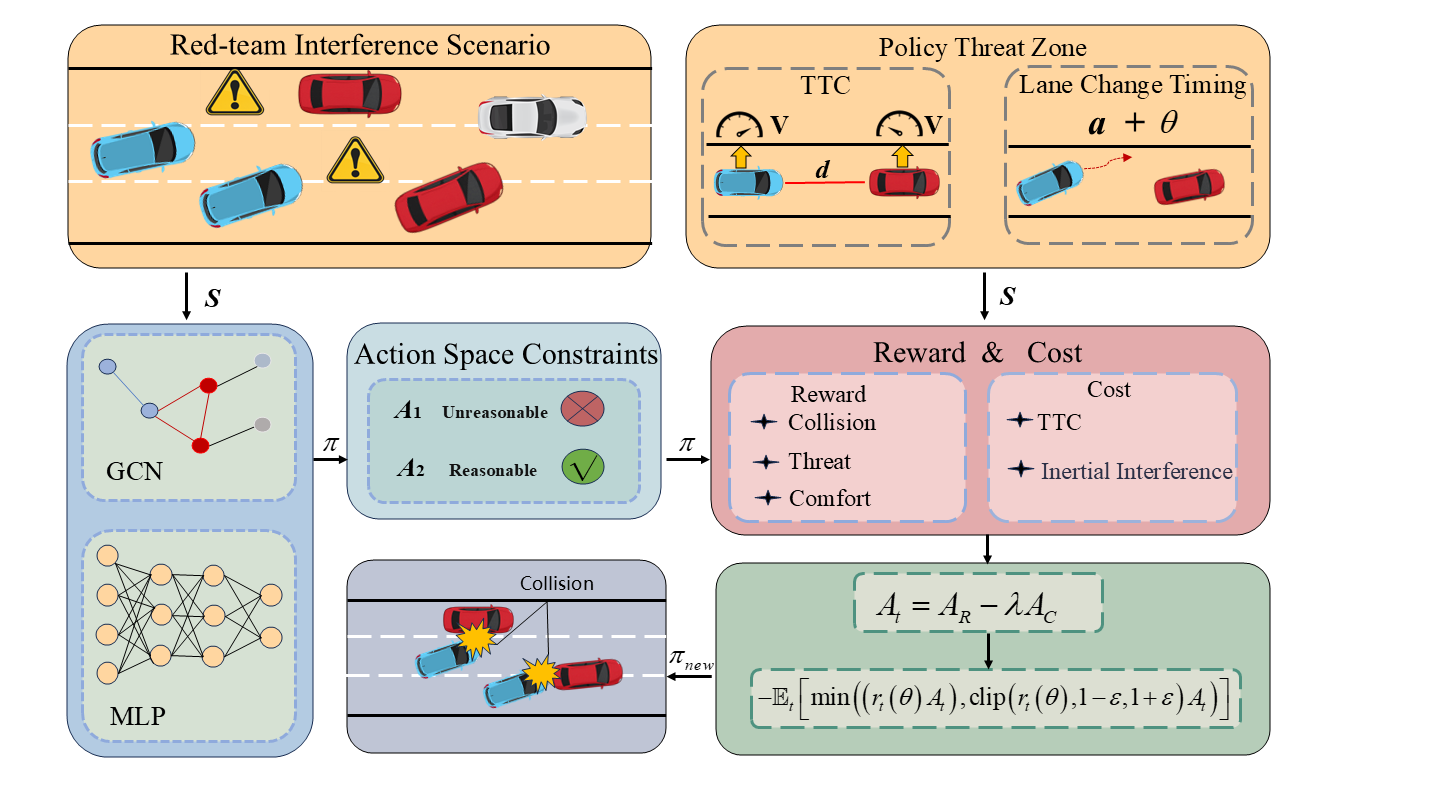}
     \caption{Overall architecture of the designed framework.}
     \label{fig:framework1}
\end{figure}

This section describes the red-team interference scenario under study. First, we represent the decision-making process of red-team vehicles  as a CGMDP. Next, we introduce a PTZ model that integrates state space information to inform the red-team high-risk action. The overall system architecture is illustrated in \cref{fig:framework1}.

\subsection{Leading Vehicle Emergency Braking}

A comprehensive analysis of the lead vehicle emergency braking scenarios is crucial for enhancing AVs’ decision-making performance. In this context, the time-to-collision (TTC) frequently falls below two seconds \cite{ttclimit}, imposing severe constraints on reaction time. Consequently, AVs must rapidly choose between emergency braking and evasive maneuvers. These rapid decisions require both real-time assessment of the lead vehicle's motion and adaptive responses to diverse traffic conditions. Therefore, a thorough research of this scenario is vital to advancing AVs decision-making performance.

\begin{figure}[htbp]
   \centering
   \subfigure[]{
        \begin{minipage}[t]{0.45\linewidth}
            \includegraphics[width=4cm]{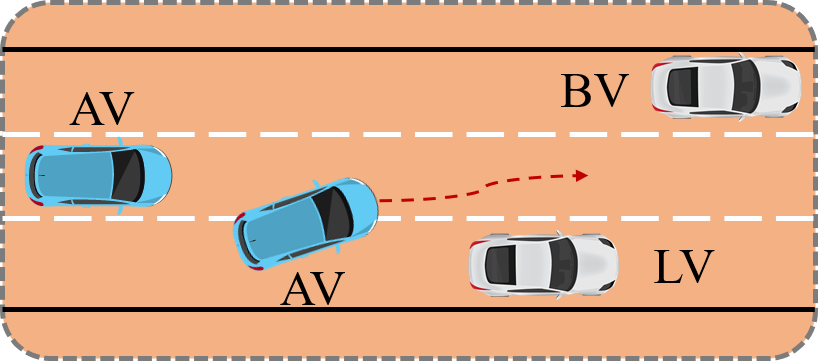}
        \end{minipage}
}
    \subfigure[] {
      \begin{minipage}[t]{0.45\linewidth}
            \includegraphics[width=4cm]{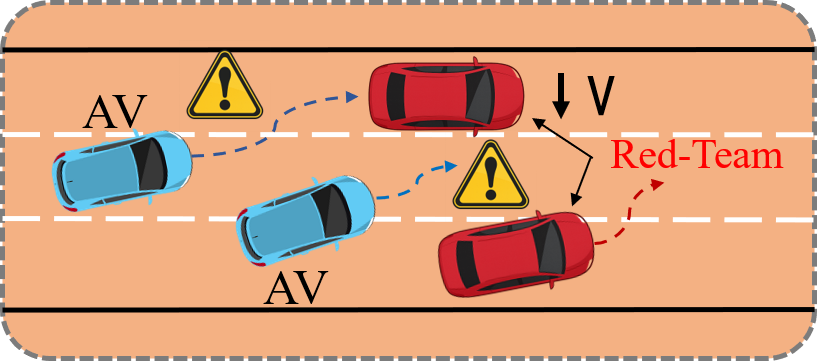}
            
        \end{minipage}
    }
    \caption{Illustration of scenarios.(a) is the lead vehicle emergency braking scenario, (b) is the red-team interference scenario.}
    
    \label{fig:scenario}
\end{figure}

In \cite{DRS}, we evaluated the AVs’ evasive policies in the lead vehicle emergency braking scenario (illustrated in Fig.~\ref{fig:scenario}(a), where AV denotes the autonomous vehicle, LV denotes the lead vehicle, and BV denotes the background vehicle) by training with DQN, D3QN, PPO, and DRS-PPO algorithms. Simulation results showed that AVs trained with DRS-PPO achieved the highest cumulative reward and reduced the collision rate to 5\%, representing a significant improvement over the other methods. However, this study considered only the LV’s solitary deceleration behavior, thereby reducing the scenario’s danger level and allowing AVs to easily discover effective counter policies. Consequently, focusing solely on simple static scenarios cannot satisfy the decision-making requirements of AVs in corner cases.

Therefore, to amplify the danger level, we extend the study in \cite{DRS} by modeling the lead vehicle and BVs as red-team agents that challenge the AVs’ optimal evasive policy. Through coordinated interference, these red-team agents maximize AVs collision rates while ensuring their actions remain compliant with traffic regulations. The red-team interference scenario is illustrated in Fig.~\ref{fig:scenario}(b).

\subsection{Constraint Graph Representation MDP}

In highly structured  traffic scenarios, the standard Markov Decision Process (MDP) \cite{mdp} fails to satisfy two critical requirements: first, it cannot enforce strict constraints on the agent's actions to avoid traffic rule violations or non-human behaviors; second, it fails to capture the complex dependencies between vehicles.

Therefore, we propose a CGMDP model based on the traditional MDP. First, we model the road network and vehicle interactions as a weighted graph $G(V, E)$, with each vehicle represented as a node $v_i \in V$ and connections between vehicles as edges $E$ that extract global state features \cite{gcn}\cite{gao2024}. Next, a constraint set $\Omega$ is applied at each decision step to eliminate actions that could violate traffic rules or result in non-human behaviors. Finally, we construct the CGMDP model as a seven-tuple:
\begin{align}
M_{CG}=(S,A,P,R,\gamma ,\Omega ,G),
\end{align}
where $S$ denotes the global state space,$A$ denotes the action space, generated by the policy model $\pi$, $P$ is  the state-transition function, $R$ is the reward function, $\gamma$ is the discount factor, $\Omega$ denotes the constraint set, $G$ is the weighted graph.

\begin{figure}[bt!]
   \centering
     \includegraphics[scale=0.3]{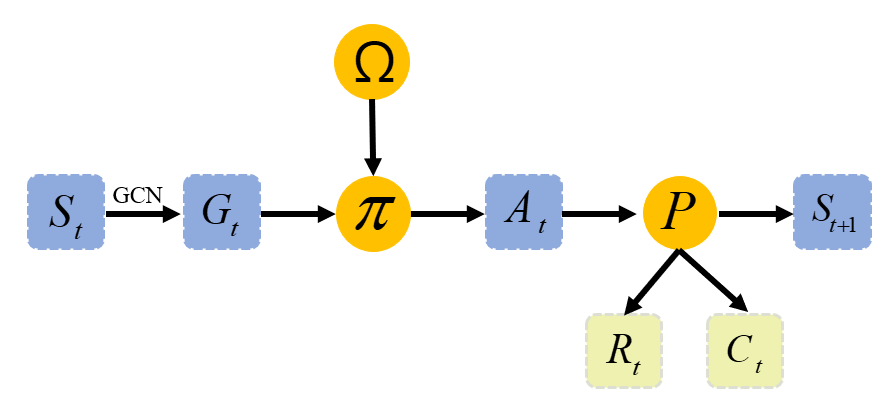}
     \caption{Constraint Graph Representation Markov Decision Process.}
     \label{fig:CGMDP}
\end{figure}

The CGMDP chain is illustrated in \cref{fig:CGMDP}, at time step $t$, the global state $S_t$ is fed into the policy model $\pi$. Leveraging the weighted graph $G$ and constraint set $\Omega$, the policy model$\pi$ selects action $A_t$ to satisfy driving objectives. The state transition function $P$ then executes action $A_t$, resulting in the next state $S_{t+1}$. Subsequently, reward $R_t$ and cost $C_t$ are calculated and employed to update the policy network.

\subsection{Policy Threat Zone Model}

In red-team interference scenarios, we develop a PTZ model to identify high-value interference targets. The state space is defined using the basic vehicle information, providing data for the policy threat zone, which serves as the basis for generating high-risk actions by the red-team vehicles.

\subsubsection{State Space}
We represent the red-team interference scenario as an undirected graph, where each vehicle corresponds to a node and each interaction to an edge. The global state space is encoded using three matrices: the node feature matrix $N_t$, the adjacency matrix $A_t$, and the mask matrix $M_t$. The construction of each matrix is founded on \cite{DRS}. Each matrix is detailed below.

Node Feature Matrix $N_t$: Each vehicle is represented by its position, heading, speed, lane index, vehicle type, and relationships with up to six nearby vehicles. The node feature matrix $N_t$ encodes these attributes for all vehicles in the scenario, as specified below:
 \begin{align}
{{N}_{t}}=\left[ {\mathbf{{x}_{i}}},{\mathbf{{y}_{i}}},{\mathbf{{\theta }_{i}}},{\mathbf{{v}_{i}}},\mathbf{{{l}_{i}}},{\mathbf{{c}_{i}}},\mathbf{{{p}_{i}}} \right],
\end{align}
where $\mathbf{x_{i}}$ and $\mathbf{y_{i}}$ denote the vehicles’ lateral and longitudinal coordinates, respectively, $\mathbf{\theta_i}$ is the heading angle, $\mathbf{v_i}$ is the speed, $\mathbf{l_i}$ is a three-dimensional one-hot encoding of the current lane, $\mathbf{c_i}$ is a one-hot encoding of the vehicle type, $\mathbf{p_i}$ encodes the relative motion relationships with up to six surrounding vehicles, Specifically, the relative relationship $p$ between each vehicle and surrounding vehicles is defined as:
\begin{align}
p=\left[ \Delta {{d}_{j}},\Delta {{v}_{j}} \right],j=\{f,r,lf,lr,rf,rr\},
\end{align}
where $\Delta {d}_{j}$ and $\Delta {v}_{j}$ denote the relative longitudinal position and relative velocity between the ego vehicle and neighbors. $f,r,lf,lr,rf,rr$ respectively represent the neighboring vehicles in front of, behind, left front of, left rear of, right front of, and right rear of the ego vehicle.

Adjacency Matrix $A_t$: The adjacency matrix represents vehicle interaction and communication among red-team vehicles, enabling information sharing. In real-world deployments, V2X communication is constrained by several practical factors. First, only a small proportion of vehicles are equipped with V2X on-board units, whereas most HVs human-driven vehicles remain silent. Second, the effective transmission range is limited by carrier frequency, transmission power, and electromagnetic interference. So, the calculation of the adjacency matrix is based on four assumptions: (a) all red-team vehicles can communicate with each other; (b) only vehicles within the perception range of red-team vehicles can exchange information; (c) human-driven vehicles cannot communicate; (d) each vehicle inherently communicates with itself.

Under the above assumptions, the adjacency matrix $A_t$ can be represented as follows:
\begin{align}
{{A}_{t}}={{\left( {{a}_{i}}_{j} \right)}_{1\le i\le n,1\le j\le n}}\in {{\mathbb{R}}^{n\times n}},
\end{align}
where $a_{ij}$ represents the connection edge between vehicle $i$ and $j$ in the graph. If ${{a}_{i}}_{j}=1$ it indicates that vehicle 
$i$ and $j$ in the graph are able to share information.

Mask Matrix $M_t$: The mask matrix filters action outputs of non-red-team vehicles and is defined as follows:
\begin{align}
{{M}_{t}}=\left[ {{m}_{1}},{{m}_{2}},\cdots ,{{m}_{i}},\cdots ,{{m}_{n}} \right],
\end{align}
where ${m}_{i}$ is defined as 1 if the ith vehicle is a red-team, and 0 otherwise.

\subsubsection{Policy Threat Zone}
To quantify the threat that red-team vehicles pose to AVs, we construct a PTZ incorporating state space information. By capturing potential risks in the scenario, the PTZ amplifies the threat to AVs. The PTZ model can be formulated as:
\begin{align}
\mathcal{D_\mathrm{PTZ}}=\bigcup\limits_{\left( ij \right)\in \Bbbk }{\left\{ \left( x,y,v,{{a}_{i}},{{\theta }_{i}} \right)|\left\{ \begin{matrix}
   {{\Phi }_{1}}(x,y,v)  \\
   {{\Phi }_{2}}\left( {{a}_{i}},{{\theta }_{i}} \right)  \\
\end{matrix} \right\} \right.},
\end{align}
where $\Bbbk$ denote the set of all AVs and red-team vehicles in the current scenario, $i$ is the subset of AVs, $j$ is the subset of red-team, $x_{i}$ and $y_{i}$ denote the vehicle’s lateral and longitudinal coordinates, $v_i$ is the speed, $a_i$ denotes AV’s acceleration, $\theta_i$ is AV’s heading angle. TTC between any two vehicles is computed from their relative positions and velocities, and serves as a metric for their degree of danger $\Phi_1$. Moreover, an AV’s acceleration and heading angle reveal whether it is executing an avoidance maneuver, thereby indicating the overall danger level $\Phi_2$  of the current scenario.

By leveraging the PTZ model, red-team vehicles can accurately identify latent risks and adjust their maneuvers to counter AVs avoidance policies, thereby intensifying scenario danger and significantly elevating the decision-making challenge for AVs. 

\subsubsection{Action Space} 
In red-team interference scenarios, each vehicle’s action vector comprises longitudinal acceleration, deceleration, and lateral steering; accordingly, the action space is defined as follows:
\begin{align}
{{\mathcal{A}}_{i}}=[{{a}_{i}},{{\delta }_{i}}],\quad {{a}_{i}}\in [-{{a}_{\text{max}}},{{a}_{\text{max}}}],{{\delta }_{i}}\in [-{{\delta }_{\text{max}}},{{\delta }_{\text{max}}}],
\end{align}
where $a$ and $\delta$ denote the vehicle’s acceleration and steering angle, respectively, while $a_{max}$ and $\delta _{max}$ represent the maximum acceleration and maximum steering angle.

To balance vehicle maneuverability and computational efficiency, we discretize longitudinal acceleration into 11 levels and steering angle into 13 levels, restrict steering to instances of zero longitudinal acceleration, and yield a 23-dimensional action space.

\section{Methodology}
In this section, building upon the PPO algorithm \cite{PPO}, we present DC-GPPO algorithm, which narrows the action space and applies a cost function to softly constrain red-team behaviors, establishing dual constraints. We further incorporate a graph convolutional network to enhance policy learning. And we also provide a detailed formulation of the reward function.

\subsection{GPPO}
PPO is a policy gradient RL algorithm that integrates importance sampling with a clipping mechanism to facilitate efficient policy iteration in complex environments. It consists of two principal components: Actor network and Critic network. The Actor network generates a stochastic policy by outputting an action probability distribution, whereas the Critic network provides state-value estimates. These estimates are combined with generalized advantage estimation (GAE)\cite{gae} to compute the advantage function at each time step, quantifying the return deviation of an action relative to the baseline policy. The policy is updated by minimizing the following loss function $L\left( \theta  \right)$.
\begin{align}
L(\theta) = -\mathbb{E}_t \Bigl[
  &\min\Bigl(
    \frac{\pi_\theta(a_t\mid s_t)}{\pi_{\theta_{\mathrm{old}}}(a_t\mid s_t)}\,A_t, \nonumber\\
  &\operatorname{clip}\Bigl(
    \frac{\pi_\theta(a_t\mid s_t)}{\pi_{\theta_{\mathrm{old}}}(a_t\mid s_t)},
    1-\varepsilon,\,1+\varepsilon
  \Bigr)\,A_t
  \Bigr)\Bigr],
\end{align}
where $A_t$ denotes the advantage function at each time step, defined as:
\begin{align}
\left\{ \begin{array}{*{35}{l}}
   {{\delta }_{t}}={{r}_{t}}+\gamma V\left( {{s}_{t}}{{_{+}}_{1}} \right)-V\left( {{s}_{t}} \right)  \\
   {{A}_{t}}=\sum\limits_{l=0}^{T-t}{{{\left( \gamma \lambda  \right)}^{l}}{{\delta }_{t}}{{_{+}}_{l}}}  \\
\end{array} \right.,
\end{align}
where ${{\delta }_{t}}$ denotes the temporal-difference (TD) error at time step $t$, $r_t$ denotes the immediate reward at time step $t$, $V\left( {{s}_{t}}{{_{+}}_{1}} \right)$ and $V\left( {{s}_{t}} \right)$ denote the state-value estimates at time steps $t$ and $t+1$, respectively, and $\gamma$ is the discount factor.

Based on the conventional PPO algorithm, we propose an enhanced GPPO. First, to keep red-team perturbations consistent with human driving, we insert an action space filter, before the policy network generates actions, the action space is constrained according to the current state of vehicles to eliminate infeasible maneuvers. Next, we integrate a Graph Convolutional Network (GCN)\cite{GCNS} to process graph-structured environmental states and extract complex spatial features. This enables the agent to capture dynamic interactions more effectively, thereby improving the efficiency of policy learning.

\subsection{DC-GPPO}
PPO-Lagrangian (PPO-Lag) \cite{ppo-lag} extends the PPO algorithm by incorporating Lagrangian multipliers into its unconstrained objective, unifying the maximization of expected cumulative rewards with the satisfaction of cost constraints within a single optimization framework. This architecture parallels the human brain’s decision-making process, optimizing performance while balancing resource utilization and safety risk.

We extend GPPO by integrating Lagrangian multipliers and a cost function into a dual-layer optimization framework, termed DC-GPPO, to enforce hard action-space constraints alongside soft behavioral constraints. DC-GPPO first prohibits actions that violate traffic regulations or cause malicious collisions, penalizing inertial interference by red-team vehicles through the cost function. Consequently, the DC-GPPO advantage $A_t^{DC}$ at time step $t$ comprises the reward advantage $A_t^{R}$ and the cost advantage $A_t^{C}$, defined respectively as follows:
\begin{align}
\left\{ \begin{array}{*{35}{l}}
   {{\delta }_{t}}^{R}={{r}_{t}}+\gamma {{V}_{R}}\left( {{s}_{t}}{{_{+}}_{1}} \right)-{{V}_{R}}\left( {{s}_{t}} \right)  \\
   {{A}_{t}}^{R}=\sum\limits_{l=0}^{T-t}{{{\left( \gamma \lambda  \right)}^{l}}{{\delta }_{t}}{{_{+}}_{l}}^{R}}  \\
\end{array} \right.,
\end{align}
\begin{align}
\left\{ \begin{array}{*{35}{l}}
   {{\delta }_{t}}^{C}={{c}_{t}}+\gamma {{V}_{C}}\left( {{s}_{t}}{{_{+}}_{1}} \right)-{{V}_{C}}\left( {{s}_{t}} \right)  \\
   {{A}_{t}}^{C}=\sum\limits_{l=0}^{T-t}{{{\left( \gamma \lambda  \right)}^{l}}{{\delta }_{t}}{{_{+}}_{l}}^{C}}  \\
\end{array} \right.,
\end{align}
where $r_t$ and $c_t$ denote the reward and cost received at time step $t$, ${{V}_{R}}\left( {{s}_{t}} \right)$ and ${{V}_{C}}\left( {{s}_{t}} \right)$ denote the corresponding state value and cost estimates at time step $t$.

By combining the reward advantage ${{A}_{t}}^{R}$ and ${{A}_{t}}^{C}$, we define the composite advantage function
${{A}_{t}}^{DC}$ as:
\begin{align}
{{A}_{t}}^{DC}={{A}_{t}}^{R}-\lambda {{A}_{t}}^{C}
\end{align}

The policy loss is then computed using this composite advantage and used to update the policy.

\subsection{Reward and Cost}
We introduce a conflict-driven danger reward function for red-team vehicles, designed to disrupt AVs decision-making. It integrates inter-vehicle distance, speed, collision events, and ride comfort to progressively induce interference behaviors \cite{iotj}\cite{itsc}. The reward function is defined as follows:
\begin{align}
\left\{ \begin{array}{*{35}{l}}
   {{R}_{\text{col}}}={{C}_{0}}, \text{if  collision}  \\
   {{R}_{\text{dan}}}={{f}_{1}}\left( d \right)  \\
   {{R}_{\text{vel}}}={{f}_{2}}\left( \Delta v \right)  \\
   {{R}_{\text{com}}}={{f}_{3}}({{a}_{\text{lat}}},{{a}_{\text{lon}}})  \\
\end{array} \right.,
\end{align}
where ${C}_{0}$ is the constant reward when an AV collides with a red-team vehicle, and $d$ denotes the relative distance between the AV and the leading red-team vehicle.The velocity reward function ${{f}_{2}}\left( \cdot  \right)$   calculates rewards based on speed change. The comfort reward function ${{f}_{3}}\left( \cdot  \right)$ evaluates ride comfort by penalizing excessive lateral acceleration $a_{\text{lat}}$ and longitudinal acceleration $a_{\text{lon}}$.

Furthermore, to prevent inertial interference from red-team vehicles, we define the corresponding cost function as follows:
\begin{align}
\left\{ \begin{array}{*{35}{l}}
   {{C}_\text{int}}={{C}_{1}}, \text{if inertial interference}  \\
   {{C}_\text{ttc}}={{f}_{4}}\left( \text{TTC} \right)  \\
\end{array} \right.,
\end{align}
where $C_1$ denotes the constant cost value assigned when a red-team vehicle exhibits inertial interference, for example, when an AV executes a lane-change maneuver, the red-team vehicle fails to intervene promptly. $\text{TTC}$ represents the time to collision between the AV and the red-team vehicle. As $\text{TTC}$ increases, the threat posed by the red-team vehicle diminishes, thereby resulting in a higher assigned cost value.

\section{Experiments}
This section presents simulation experiments designed to evaluate the proposed framework. We compare four RL algorithms: Graph-D3QN (GD3QN)\cite{sensors}\cite{rate}, PPO\cite{PPO}, GPPO and DC-GPPO. And we conducted an analysis from two aspects: the training results and the testing results.

\subsection{Simulation Setup}
We use the SUMO simulator \cite{sumo} to create a red-team interference scenario featuring single-vehicle interference (SVI) and multi-vehicle interference (MVI). AVs use the decision-making model from our previous work \cite{DRS}, red-team vehicles follow our proposed algorithms, and all other BVs rely on SUMO’s default driver model. We train red-team agents using four RL algorithms—PPO, GD3QN, GPPO, and DC-GPPO—to disrupt AVs behavior. Experimental hyperparameters and simulation settings are presented in \cref{tab:hyperparams}.

\begin{table}[htbp]
  \centering
   \captionsetup{
    font=small,
    singlelinecheck=off, 
    justification=centering 
  }
  \caption{THE HYPERPARAMETERS AND SIMULATION ENVIRONMENT PARAMETERS SETTING}
  \label{tab:hyperparams}
  \begin{tabular}{@{} c c @{}}
    \hline
    \textbf{Parameter}            & \textbf{Value}  \\          
    \hline
    Number of training episodes & 500               \\
    Batch size                & 32                  \\
    Discount factor           & 0.9                 \\
    Learning rate             & 7.5$\times10^{-4}$  
    \\
    Starting greedy rate      & 0.8                 \\
    Ending greedy rate        & 0.05                \\
    Optimizer                 & Adam                \\
    Nonlinearity              & ReLU \cite{relu}                \\
    Time step                 & 0.1\,s              \\
    Max acceleration          & 5$\,\mathrm{m/s^2}$ \\
    \hline
  \end{tabular}
\end{table}

\subsection{Training Results}

\begin{figure}[htbp]
   \centering
   \subfigure[]{
        \begin{minipage}[t]{0.45\linewidth}
            \includegraphics[width=4cm]{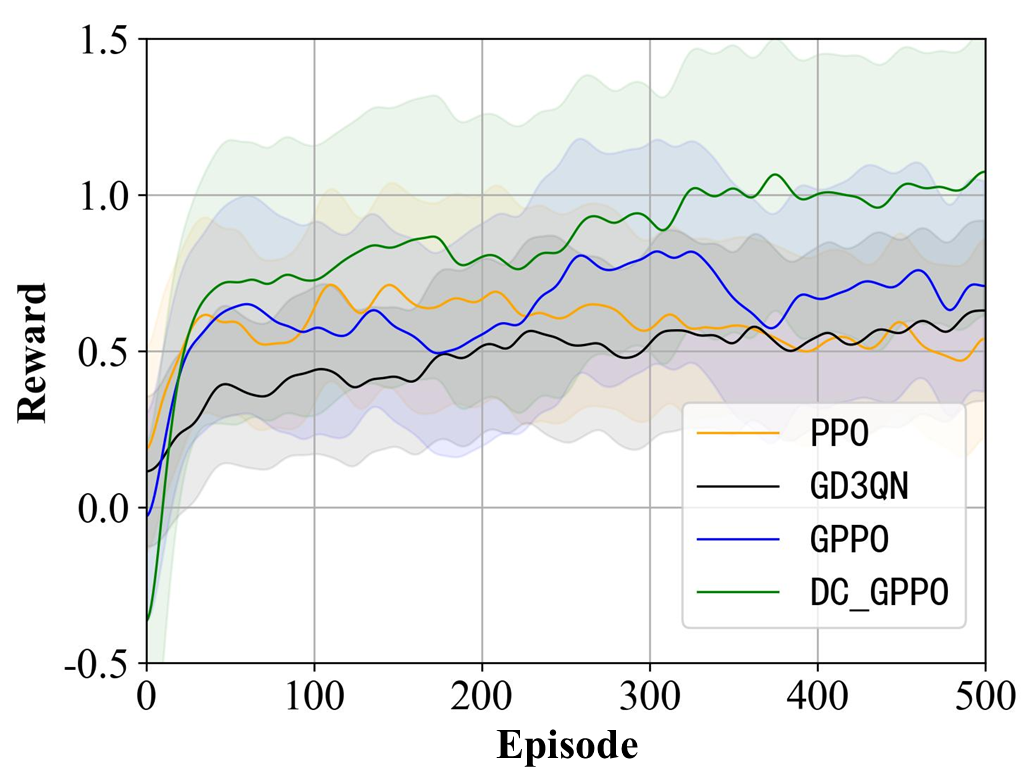}
        \end{minipage}
}
    \subfigure[] {
      \begin{minipage}[t]{0.45\linewidth}
            \includegraphics[width=4cm]{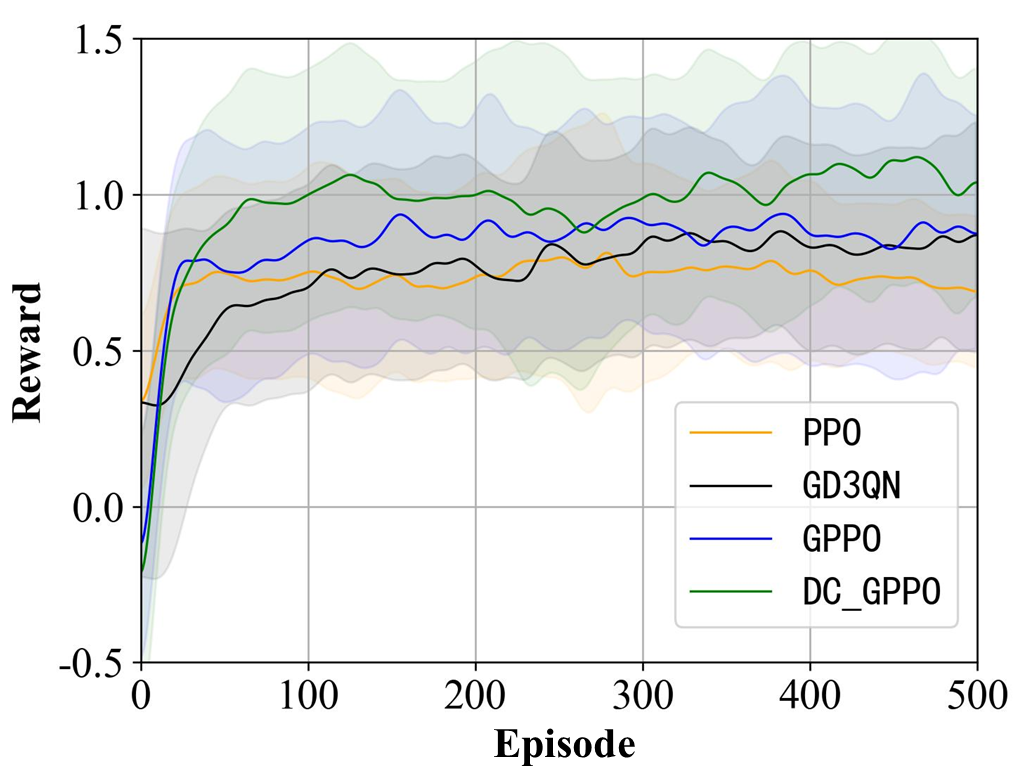}
            
        \end{minipage}
    }
    \caption{Illustration of reward curves.(a) is the single-vehicle interference scenario, (b) is the multi-vehicles interference scenario. The shaded areas show the standard deviation for 5 random seeds.}
    
    \label{fig:reward curve}
\end{figure}

\cref{fig:reward curve} present the reward curves of four algorithms in SVI and MVI. We employ the normalized reward \cite{liu2024mv} metric to rigorously assess overall algorithm performance. Higher reward values indicate that the red-team poses a greater threat to AVs, reflecting a more dangerous scenario and an increased probability of AVs decision-making failure. In both SVI and MVI, the DC-GPPO algorithm’s reward curve markedly exceeds those of the other methods, indicating that the red-team inflicts the greatest threat on AVs and substantially elevates the scenario’s degree of danger.

\subsection{Test Results}

First, without red-team interference, we tested AVs subjected solely to BVs performing emergency braking. These results established the baseline for subsequent comparisons. Next, we assessed the AVs under red-team interference, where the red-team was trained with the four algorithms described above, employing the same metrics. The comparative results are presented in \cref{tab:performance_comparison}. In \cite{DRS}, without interference, the AV’s collision rate was only 5\%, but under red-team interference it rose to 85\%, indicating a substantial degradation of AVs decision-making safety. Moreover, collisions under interference shortened the AV’s travel time. The AV’s average lateral acceleration and average speed reflect the red-team’s lateral and longitudinal perturbations: average lateral acceleration increased markedly from $0.10 m/s^2$ to $0.34 m/s^2$, whereas longitudinal impacts remained negligible.

\begin{table}[htbp]
\centering
\caption{TESTING COMPARATIVE RESULTS.}
\label{tab:performance_comparison}

\begin{tabular}{lcccccc}
\toprule
\textbf{} & \textbf{Metric} & \textbf{BL} & \textbf{PPO} & \textbf{GD3QN} & \textbf{GPPO} & \textbf{DC-GPPO} \\
\midrule
\multirow{4}{*}{SVI} 
& CR & 5.00 & 37.00 & 50.00 & 52.00 & \textbf{75.00}\\
& TT & 10.86 & \textbf{9.02} & 9.05 & 9.10 & 9.06 \\
& ALA & 0.11 & \textbf{0.18} & 0.17 & 0.16 & 0.16 \\
& AS & \textbf{14.18} & 15.61 & 15.53 & 15.92 & 15.66 \\
\midrule
\multirow{4}{*}{MVI} 
& CR & / & 67.00 & 55.00 & 70.00 & \textbf{85.00} \\
& TT & / & 8.84 & 8.72 & 8.41 & \textbf{7.97} \\
& ALA & / & 0.21 & 0.28 & 0.26 & \textbf{0.34} \\
& AS & / & 14.93 & 14.68 & 14.51 & \textbf{13.96}\\
\bottomrule
\end{tabular}
\begin{tablenotes}
\item BL:Baseline, CR: Collision rate (\%), TT: Travel time ($\mathrm{s}$), ALA: Avg. lateral acceleration (${\rm{m/s^{2}}}$), AS: Average speed (${\rm{m/s}}$).
\end{tablenotes}
\end{table}

\begin{figure*}[bt!]
    \centering
    \subfigure[]{
        \begin{minipage}[t]{0.23\linewidth}
            \includegraphics[height=2.5cm]{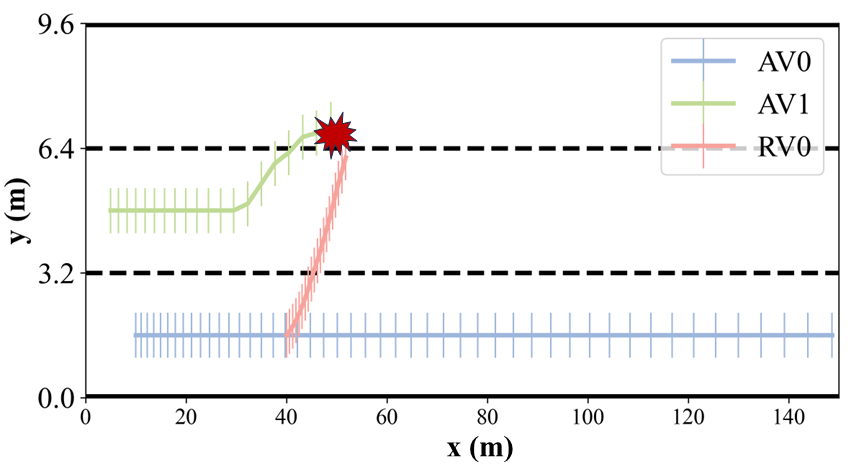}
            
        \end{minipage}
    } 
    \subfigure[]{
        \begin{minipage}[t]{0.23\linewidth}
            \includegraphics[height=2.5cm]{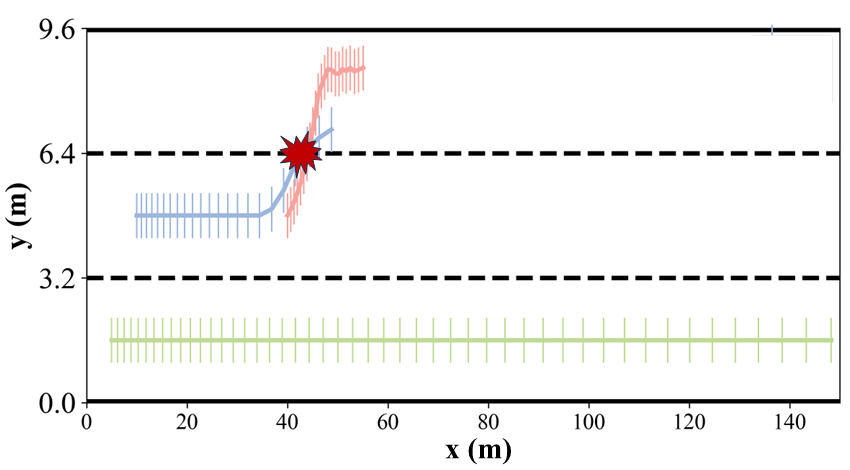}
            
        \end{minipage}
    }
    \subfigure[]{
        \begin{minipage}[t]{0.23\linewidth}
            \includegraphics[height=2.5cm]{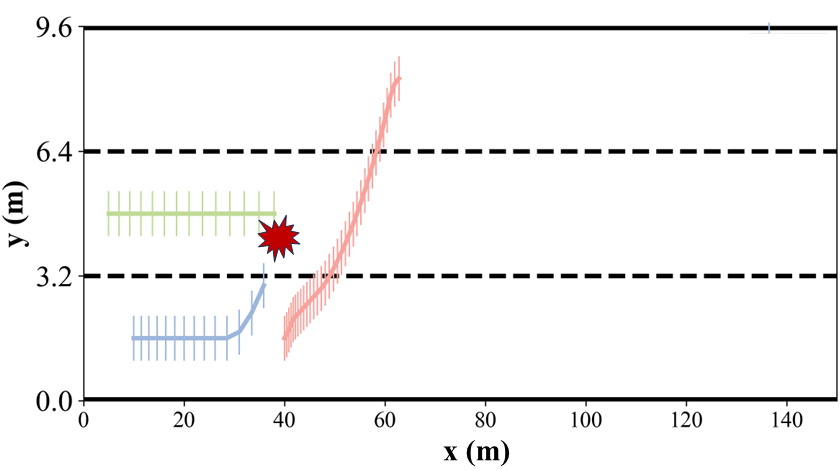}
            
        \end{minipage}
    }
    \subfigure[]{
        \begin{minipage}[t]{0.23\linewidth}
            \includegraphics[height=2.5cm]{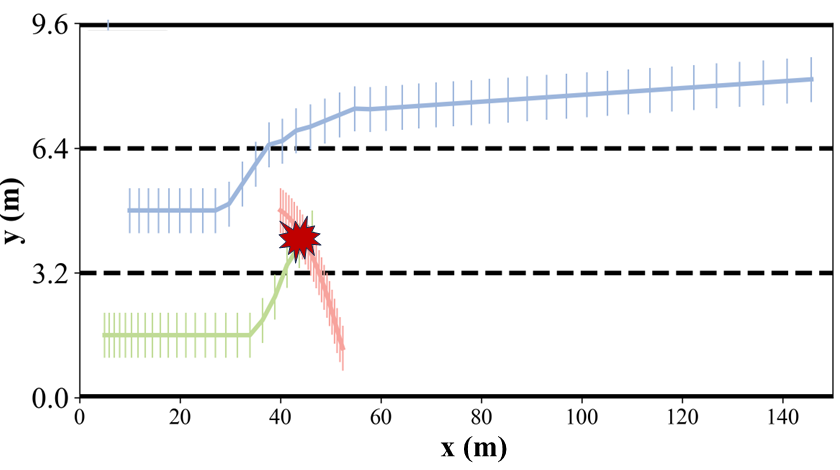}
        \end{minipage}
    }

  \vspace{0.5ex} 

    \subfigure[]{
        \begin{minipage}[t]{0.23\linewidth}
            \includegraphics[height=2.5cm]{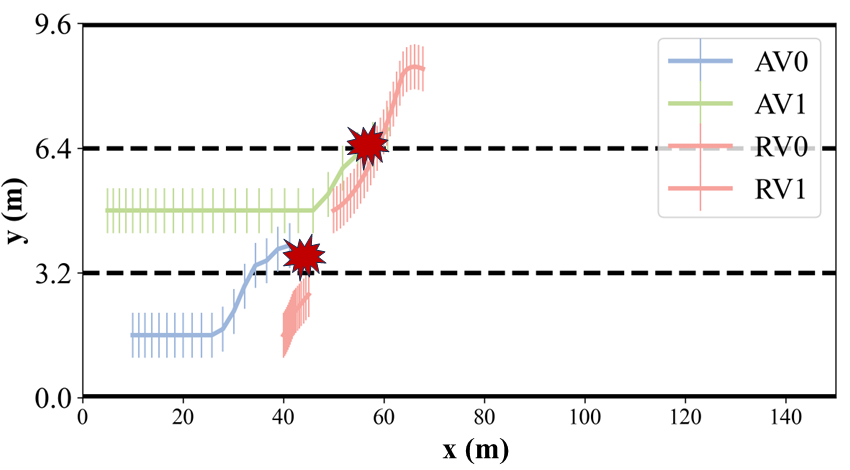}
        \end{minipage}
  }
   \subfigure[]{
        \begin{minipage}[t]{0.23\linewidth}
            \includegraphics[height=2.5cm]{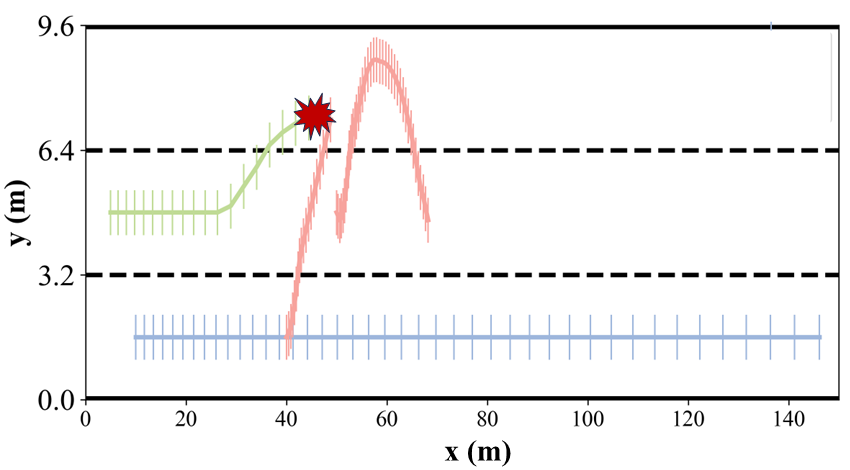}
        \end{minipage}
  }
   \subfigure[]{
        \begin{minipage}[t]{0.23\linewidth}
            \includegraphics[height=2.5cm]{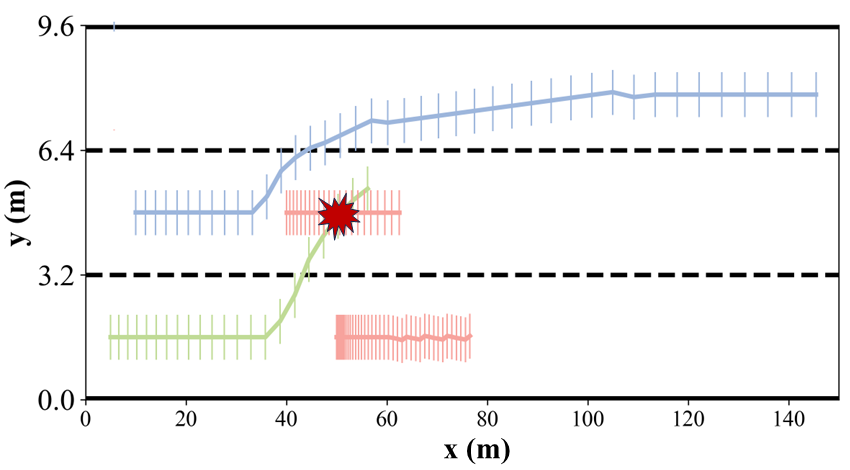}
        \end{minipage}
  }
   \subfigure[]{
        \begin{minipage}[t]{0.23\linewidth}
            \includegraphics[height=2.5cm]{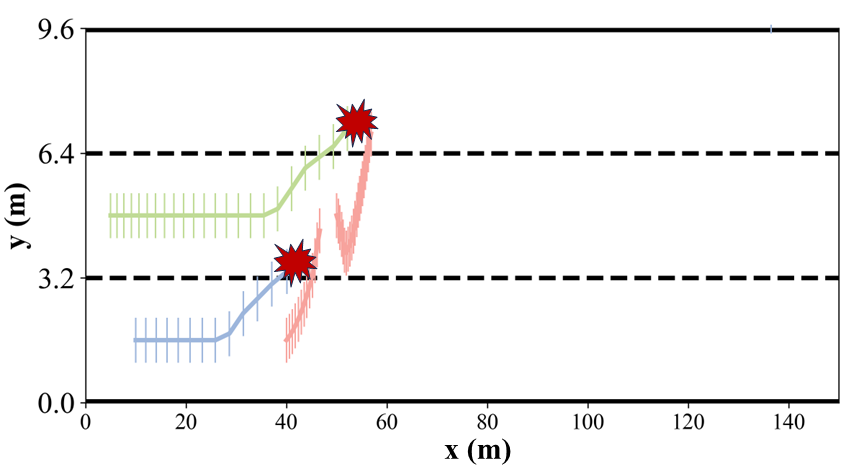}
        \end{minipage}
  }
    \caption{Trajectories of red-team vehicles and AVs. AV0 and AV1 represent the trajectories of AVs, while RV0 and RV1 represent the trajectories of red-team vehicles. (a)-(d) illustrate the trajectories of AVs and red-team vehicles under SVI, (e)-(h) illustrate the trajectories of AVs and all red-team vehicles under MVI, with markers indicating collision locations. }
    \label{fig:trajs}
\end{figure*}

\cref{fig:trajs} presents the trajectories of red-team vehicles and AVs during testing. As shown, red-team vehicle exploration led to the emergence of multiple corner cases, which  AVs could not handle, resulting in collisions.

In summary, we observe that all four algorithms impact the decision-making of AVs and contribute to an increase in the overall danger of the scenarios. This further validates the effectiveness of the proposed framework. Through the red-team vehicle's exploration, more corner cases are successfully identified, thereby presenting greater challenges to AVs decision-making in safety-critical scenarios.

\section{CONCLUSIONS}
Existing research on safety-critical scenarios lacks exploration of corner cases. We propose a RMARL framework. In this framework, BVs are redefined as red-team vehicles with interference capabilities. Their decision-making process is modeled as CGMDP. By incorporating hard constraints on the action space and soft behavioral constraints, the red-team vehicles ensure compliance with safety regulations while disrupting the AVs. The PTZ model quantifies the threat posed by red-team vehicles to AVs, encouraging them to take more extreme actions. The results show that the RMARL framework overcomes the limitations of traditional static scenarios, actively exploring corner cases beyond the data distribution, and providing a new direction for safety-critical scenario research.

In future research, we will broaden the framework from an emergency braking scenario to a wider set of safety-critical scenarios, including multi-lane highways and complex intersections. The approach will also be evaluated with higher-fidelity simulators. Furthermore, to meet the fine-grained control requirements of safety-critical decision making, we will replace the discrete action space with a continuous one, enabling smoother and more precise maneuvers. Additionally, we will optimize the efficiency of red-team strategy generation and adversarial training, enhancing the specificity and stealth of interference behaviors.

\bibliographystyle{ieeetr}
\small\bibliography{reference}

\end{document}